\documentclass[journal]{IEEEtran}
\usepackage[english]{babel}
\usepackage[utf8]{inputenc}
\usepackage{listings}% http://ctan.org/pkg/listings
\usepackage{caption}
\usepackage{DejaVuSans}
\usepackage{DejaVuSansMono}
\usepackage{amssymb}
\usepackage{amsmath}

\usepackage{subcaption}
\usepackage{graphicx}

\usepackage{alltt}

\usepackage{fancyvrb}

\usepackage[table,xcdraw]{xcolor}
\usepackage{multirow}

\usepackage[T1]{fontenc}
\usepackage{fancyvrb}
\usepackage{hyperref}
\usepackage{listings}
\usepackage{gensymb}
\usepackage[scaled=.55]{beramono}    
\fvset{baselinestretch=0.5}
\let\conjugatet\overline
\DeclareCaptionFormat{listing}{\rule{\dimexpr0.9\columnwidth+17pt\relax}{0.35pt}\par\vskip1pt#1#2#3}
\IEEEoverridecommandlockouts                              % This command is only needed if 
\usepackage{blindtext}
\usepackage{graphicx}
\usepackage{listings}% http://ctan.org/pkg/listings
\usepackage{xcolor}

\usepackage{hyperref}
\hypersetup{
    colorlinks=true,
    linkcolor=blue,
    filecolor=magenta,      
    urlcolor=cyan,
    citecolor=blue,
}

\usepackage[normalem]{ulem}
\begin{document}
\title{gym-gazebo2, a toolkit for reinforcement learning using ROS 2 and Gazebo }

\author{Nestor Gonzalez Lopez, Yue Leire Erro Nuin, Elias Barba Moral, Lander Usategui San Juan,\\
Alejandro Solano Rueda, Víctor Mayoral Vilches and Risto Kojcev\\
\emph{Acutronic Robotics, March 2019}
}

\maketitle

% \thispagestyle{fancy} 			% Enabling the custom headers/footers for the first page 
% The first character should be within \initial{}

%inspired by https://www.xilinx.com/support/documentation/data_sheets/ds190-Zynq-7000-Overview.pdf

\footnotesize
\def\arraystretch{1.3}%  1 is the default, change whatever you need

\begin{abstract}
%\boldmath
This paper presents an upgraded, real world application oriented version of gym-gazebo, the Robot Operating System (ROS) and Gazebo based Reinforcement Learning (RL) toolkit, which complies with OpenAI Gym. The content discusses the new ROS 2 based software architecture and summarizes the results obtained using Proximal Policy Optimization (PPO). Ultimately, the output of this work presents a benchmarking system for robotics that allows different techniques and algorithms to be compared using the same virtual conditions. We have evaluated environments with different levels of complexity of the Modular Articulated Robotic Arm (MARA), reaching accuracies in the millimeter scale. The converged results show the feasibility and usefulness of the gym-gazebo 2 toolkit, its potential and applicability in industrial use cases, using modular robots.
\end{abstract}

\section{Introduction}
% why we come up with gym gazebo and why we upgraded to gym_gazebo 2. What is the gap we are filling?
% Brief explanation of what is gym-gazebo2.
% Motivation
% Why we created a new version.
% Our target / focus (slightly different from gym-gazebo 1).
% In gym-gazebo 1 we focused in the Turtlebot, now we focused in MARA for xyz reasons.

% CRITICS to gym-gazebo 1
% https://upcommons.upc.edu/handle/2117/113948
% [ZLVC16] aimed to provide a toolkit for UAV researchers to compare their techniques in a well-defined controlled environment. Hereby, the Gazebo/ROS setup is extended by OpenAI Gym [BCP+16]. OpenAI Gym is a toolkit for reinforcement learning research that allows researchers to have a common framework to compare and conveniently test reinforcement learning algorithms. In practice the implementation failed to be modular enough for individual needs, since the connection to OpenAI Gym is specifically configured for their selection of example environments. The missing plug-and-play nature serves rather niche use cases  and the benchmark issue remains

Gym-gazebo \cite{zamora2016extending} proves the feasibility of using the Robot Operating System (ROS) \cite{Quigley09} and Gazebo \cite{koenig2004design} to train robots via Reinforcement Learning (RL) \cite{vilches2018robot_gym, kojcev2018hierarchical, tongloy2017asynchronous, yoonpath, zenglearning, sarkar2018sequential, kim2018deep, linteaching, zhang2018danger}. The first version was a successful proof of concept which is being used by multiple research laboratories and many users of the robotics community. Given the positive impact of the previous version, specially its usability, in gym-gazebo2 we are aiming to advance our RL methods to be applicable in real tasks. This is the logical evolution towards our initial goal: to bring RL methods into robotics at a professional/industrial level. For this reason we have focused on Modular Articulated Robotic Arm (MARA), a truly modular, extensible and scalable robotic arm\footnote{\url{https://acutronicrobotics.com/products/mara/}}.

We research how RL can be used instead of traditional path planning techniques. We aim to train behaviours that can be applied in complex dynamic environments, which resemble the new demands of agile production and human robot collaboration scenarios. Achieving this would lead to faster and easier development of robotic applications and moving the RL techniques from a research setting to a production environment. Gym-gazebo2 is a step forward in this long term goal.

\section{State of the Art}
% what others have done and why we are different(better)
% AI in robotics, SoA things..
% Gym-gazebo 1, explain the base.
% Whats happening with the old version, gym-gazebo 1.
RL algorithms are actively being developed and tested in robotics \cite{2018arXiv180204082M}, but despite achieving positive results, one of the major difficulties in the application of algorithms, sample complexity, still remains. Acquiring large amounts of sampling data can be computationally expensive and time-consuming \cite{arulkumaran2017brief}. This is why accelerating the training process is a must. The most common procedure is to train in a simulated environment and then transfer the obtained knowledge to the real robot \cite{christiano2016transfer, james2017transferring, peng2018sim}. 

In order to avoid having specific use case environment and algorithm implementations, non-profit AI research companies, such as OpenAI, have created a generic set of algorithm and environment interfaces. In OpenAI's Gym \cite{1606.01540}, agent-state combinations encapsulate information in environments, which will be able to make use of all the available algorithms and tools. This abstraction allows an easier implementation and tune of the RL algorithms, but most importantly, it creates the possibility of using any kind of virtual agent. This includes robotics, which Gym is already supporting with several environments on their roster.

By following this approach, once we learn an optimal policy in the virtual replica of our real environment, we face the issue of transferring efficiently the learned behaviour into the real robot. Virtual replicas must provide accurate simulations, resembling real life conditions as much as possible. Mujoco environments \cite{todorov2012mujoco} for instance, provide the required accurate physics and robot simulations. A successful example is Learning Dexterous In-Hand Manipulation \cite{1808.00177}, where a human-like robot hand learns to solve the object reorientation task entirely in simulation without any human input. After the training phase, the learned policy is successfully transferred to the real robot. However, the process of overcoming the reality gap caused by the virtual version being a coarse approximation of the real world is a really complex task. Techniques like distributing the training over simulations with different characteristics increase the difficulty of this approach. In addition, Mujoco is locked in proprietary software, which greatly limits its use.
%\risto{this is little bit dangerous, if people are able to do sim to real transfer, why do they need gym gazebo? I would add that it was not so straightforward, according to the paper to transfer this, and that there were many differences between the control of the real system and simulated one. Second I would emphasize that Mujoco is locked in proprietary software, not open source.}

Having the goal of transferring our policies to industrial environments, a more convenient approach is to use the same development tools used by roboticists. gym\_gazebo extends OpenAI Gym focusing on utilizing the most common tools used in robotics such as ROS and the Gazebo simulator. ROS 2 has recently gained popularity due to its promising middle ware protocol and advanced architecture. MARA, our collaborative modular robotic arm, already runs ROS 2 in each actuator, sensor or any other representative module. In order to leverage its architecture and provide advanced capabilities tailored for industrial applications, we integrated the ROS 2 functionality in the new version of gym\_gazebo, gym\_gazebo2. 

The previous version of gym-gazebo had a few drawbacks. Aside from migrating the toolkit to the standard robotic middleware of the next years (ROS 2), we also needed to address various issues regarding the software architecture of gym-gazebo. The inconvenient structure of the original version caused multiple installation and usage issues for many users. Since the much needed improvements would change multiple core concepts, we decided to create a completely new toolkit, instead of just updating the previous. In gym-gazebo2, we implemented a more convenient and easy-to-tune robot specific architecture, which is simpler to follow/replicate for users wanting to add their own robots to the toolkit. The new design relies on the new Python ROS Client Library developed for ROS 2 for some new key aspects, like the launch process or initialization of a training.

\section{Architecture}

The new architecture of the toolkit consists of three main software blocks: gym-gazebo2, ROS 2 and Gazebo. The gym-gazebo2 module takes care of creating environments and registering them in OpenAI's Gym. We created the original gym-gazebo as an extension to the Gym, as it perfectly suited our needs at the time. However, although the new version we are presenting still uses the Gym to register its environments, it is not a fork anymore, but a standalone tool. We keep all the benefits provided by Gym as we make use of it as a library, but we gain much more flexibility by not having to rely on the same file structure. This move also eliminates the need of manually merging the latest updates from the parent repository from time to time.

Our agent and status specific environments interact with the robot via ROS 2, which is the middleware that allows the communication between gym-gazebo and the robot. The robot can be either a real robot or a simulated replica of the real one. As we have already mentioned, we train the robot in a simulated environment to later translate a safe and optimized policy to the real version. The simulator we are using is Gazebo, which provides a robust physics engine, high-quality graphics and convenient programmatic and graphical interfaces. But more importantly, it provides the necessary interfaces (Gazebo specific ROS 2 packages) required to simulate a robot in the Gazebo via ROS 2 using ROS messages, ROS services and dynamic reconfigure.

Unlike in the original version of this toolkit, this time we have decided to remove all robot specific assets (launch files, robot description files etc.) from the module. As we wanted to comply with the company's modularity philosophy, we took the decision of leaving all robot specific content encapsulated in packages particular for the robot properties, such as its kinematics. We only kept the robot specific environment file, which has to be added to the collection of environments of this toolkit.

Our environment files comply with OpenAI's Gym and extend the API with basic core functions, which are always agent-state specific. These basic core functions will be called from the RL algorithm. The functions that provide this interaction with the environment (and through the environment, with ROS 2) are the following:

\begin{itemize}
\item \textbf{init}: Class constructor, used to initialize the environment. This includes launching Gazebo and all the robot specific ROS 2 nodes in a separate thread.
\item \textbf{step}: Executes one action. After the action is executed, the function returns the reward obtained from the observation of the new state. This observation is returned as well, and also a Boolean field which indicates the success of that action or the end of the episode.
\item \textbf{reset}: Resets the environment by setting the robot to a initial-like state. This is easily achieved by resetting the Gazebo simulation, but it is not required to be done this way.
\end{itemize}

Since the different environments share a lot of code, we decided to create a small internal utility-API, which is only called within gym-gazebo2 environments. We have created a /utils folder inside the gym-gazebo2 module and organized the code in different files. For now, the code is classified in the following groups: 
\begin{itemize}
\item \textit{ut\_gazebo}: Utilities only related to gazebo.
\item \textit{ut\_generic}: Utilities not related to any other group.
\item \textit{ut\_launch}: Functions involved in the initialization process of an environment.
\item \textit{ut\_mara}: Utilities specific to MARA robot's environment, shared between all MARA's environments.
\item \textit{ut\_math}: Functions used to perform mathematical calculations.
\end{itemize}

\subsection{Installation}
Installation wise, we wanted to simplify the complex setup process as much as possible, but also let the user be in control of what is being executed every time. We are not using any automatic installation script since we want the user to be aware of each of the steps that need to be followed. This will lead to an easier and faster error tracking, which will facilitate providing assistance to the non Linux/ROS experienced part of the community.

We are also providing a gym-gazebo2 ready \textbf{docker container}, which will simplify the installation and avoid user's machine libraries to interfere with gym-gazebo2. In the near future, this should be the default installation option, leaving the step by step installation only to advanced users that need to create complex behaviours or add new robots to gym-gazebo2. 

\subsection{Command-line customization}
Every MARA environment provides three command-line customization arguments. You can read the details by using the -h option in any MARA script (e.g: \textit{python3 gg\_random.py -h}). The help message at release date is the following:

\begin{lstlisting}[language=bash, linewidth=\columnwidth, breaklines=true]
usage: gg_random.py [-h] [-g] [-r] [-v VELOCITY] [-m | -p PORT]

MARA environment argument provider.

optional arguments:
  -h, --help            Show this help message and exit.
  -g, --gzclient        Run user interface.
  -r, --real_speed      Execute the simulation in real speed and using the running specific driver.
  -v, --velocity        Set servo motor velocity. Keep < 1,57 for real speed. Applies only with -r --real_speed option.
  -m, --multi_instance  Provide network segmentation to allow multiple
                        instances.
  -p PORT, --port PORT  Provide exact port to the network segmentation to
                        allow multiple instances.
\end{lstlisting}

The same environment is used to train and run the learned policy, but the driver that interacts with the simulation is not the same. We use a training specific gazebo plugin (driver) that aims to achieve the maximum possible simulation speed. This training optimized driver is the default option.

Once the robot manages to learn an interesting policy, we might want to test it in a real scenario. In order to achieve this, we need a different driver that provides velocity control and is able to execute smoother actions via interpolation. We need to select the -r --real\_speed flag for this and we also might want to tune MARA's servo velocity with -v --velocity. Recommended velocities are the same as a real MARA would accept, which range from to 0 $rad/s$ to 1,57 $rad/s$.

\section{MARA Environments} 
\label{environments}

We are presenting four environments with the release of gym-gazebo2, with a plan to extend to more environments over time. We have focused on MARA first for being this modular robot arm an Acutronic Robotic product and for being the most direct option of transferring policies learned in gym-gazebo2 to the real world, hopefully industrial applications.

MARA is a collaborative robotic arm with ROS 2 in each actuator, sensor or any other representative module. Each module has native ROS 2 support and delivers industrial-grade features including synchronization, deterministic communication latency, a ROS 2 software and hardware component life-cycle, and more. Altogether, MARA empowers new possibilities and applications in the professional landscape of robotics.

\begin{center}
\captionsetup{type=figure}
  \includegraphics[height=60mm]{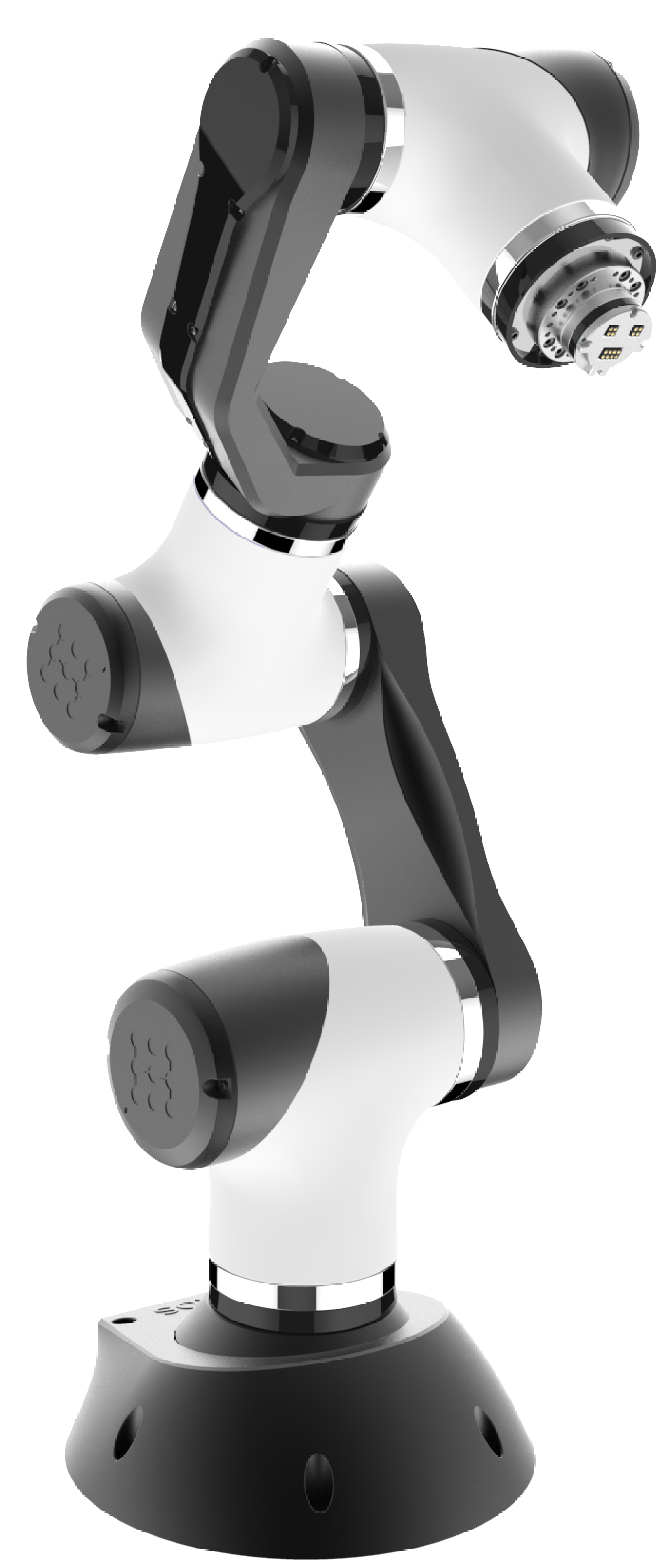}
  \caption{Real MARA, Modular Articulated Robotic Arm.}
\end{center}

In gym-gazebo2 we will be training the simulation version of MARA, which will allow to rapidly test RL algorithms in a safe way. The base environment is a 6 degrees of freedom (DoF) MARA robot placed in the middle of a table. The goal is to reach a target, which is a point in the 3D space.

\begin{center}
\captionsetup{type=figure}
  \includegraphics[height=80mm]{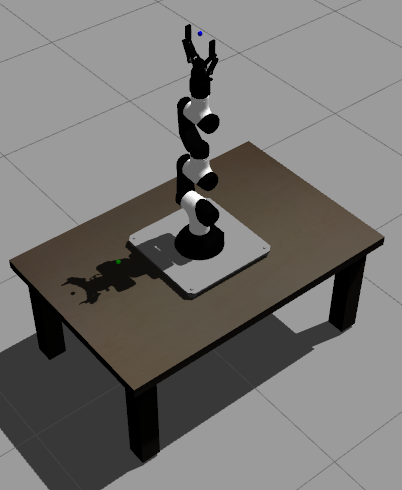}
  \caption{MARA environment: 6DoF MARA robot placed in the middle of the table in its initial pose. The green dot is the target position that the blue point (end-effector) must reach.} 
\end{center}

The following are the four environments currently available
for MARA:

\begin{itemize}
\item MARA
\item MARA Orient
\item MARA Collision
\item MARA Collision Orient
\end{itemize}

\subsection{MARA}
This is the simplest environment in the list above. We reward the agent only based on the distance from the gripper's center to the target position. We reset the environment when a collision occurs, but we do not model its occurrence into the reward function. The orientation is also omitted.

\textbf{Reward system}: The reward is calculated using the distance between the target (defined by the user), and the position of the end-effector of the robot taken from the last observation, after executing the action in that particular step. The actual formula we use is:
\begin{equation}
x = \sqrt{\frac{1}{3}\sum^{3}_{i=1}(loc_{robot}[i]-loc_{target}[i])^{2}}
\end{equation}

where $loc_{robot}$ are the Cartesian coordinates of the end-effector of the robot and $loc_{target}$ are the Cartesian coordinates of the target. Knowing this, the reward function is:

\begin{equation}
rew = \underset{rew_{dist}}{\underbrace{\frac{e^{-\alpha x}-e^{-\alpha} + 10 (e^{-\alpha \frac{x}{done}} - e^{-\alpha})} {1-e^{-\alpha}}}} - 1 
\end{equation}

where the $\alpha$ and $done$ are hyperparameters and the reward function values range from -1 to 10. This function will take negative exponential dependence with the distance between the robot and the target, and values close to 0 when close to the desired point. The part $10 (e^{-\alpha\frac{x}{done}}-e^{-\alpha})$ is meant to become important when $x\approx 0$, so $done$ must be small. It could be interpreted as the distance we would consider as a good convergence point.

\subsection{MARA Orient}
This environment takes into account the translation as well as the rotation of the end-effector of the robot. It also gets reset to the start pose when a collision takes place, but again, there is not direct impact of this type of actions in the reward.

\begin{figure*}
    \centering
        \includegraphics[width=\textwidth]{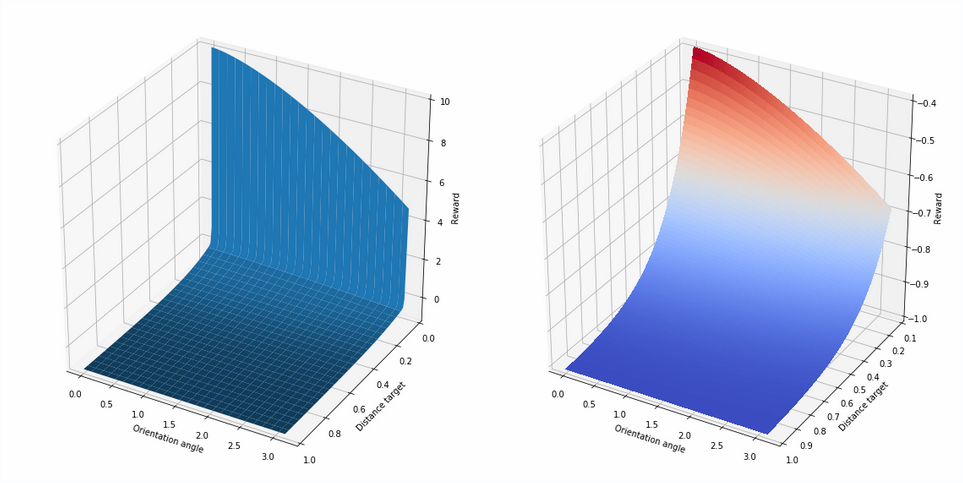}
    \caption{Core reward functional shape (Eq.\ref{eq:rew_core}) for $ \mathbf{\alpha} = 5$, $ \mathbf{\beta} = 1.5$, $ \mathbf{\gamma} = 1$, $ \mathbf{\delta} = 3$, $ \mathbf{\eta} = 0.03$, $done = 0.02$. The figure on the left is the core reward for all the range of orientation angle and target distance. The figure on the right is a closer look at the region between $0.1 \leq rew_{dist}\leq 1$, note the range change in the reward axis. Both figures show that the preferred axis of improvement is the distance target, but also that for lower values of target distance, in order to get a good reward, a good level orientation is required.}
    \label{fig:reward_shape}
\end{figure*}

\textbf{Reward system}: The reward is calculated by combining the difference of the position and orientation between the real goal and the end-effector. The distance reward is computed in the same way as in the previous environment (MARA). The difference here is the addition of an orientation reward term. In order to estimate the difference between two different poses, we use the following metric:

\begin{equation}
y = 2cos^{-1}\left(|quat_{robot}[i]*\conjugatet{quat_{target}[i]}|\right)
\end{equation}
where $quat_{robot}$ is the orientation of the effector in quaternion form and $quat_{target}$. We incorporate this to the previous formula, as a regulator of the values obtained in the distance part:

\begin{equation}
rew = \underset{rew_{core}}{\underbrace{rew_{dist}*\frac{1 + \gamma - (\frac{y}{\pi})^{\beta}}{1 + \gamma} -1}}
\label{eq:rew_core}
\end{equation}
where $reward_{dist}$ is the same fraction as in MARA environment, and $\beta$ and $\gamma$ are hyperparameters. This new term should add a penalty on having bad orientation, specially when the reward distance becomes more important. Our choice of parameters makes this term not dominant. Fig.\ref{fig:reward_shape} shows the shape of this function.

\subsection{MARA Collision}
This environment considers both the translation of the end-effector and the collisions. If the robot executes a colliding action it will get a punishment and also the reset to the initial pose. In this environment orientation is not taken into account.

\textbf{Reward system}: The reward is computed in similar manner as in MARA environment if there is no collision. Otherwise, if it gets collided, the reward is complemented with a penalty term, that depends on the reward obtained by the distance. In other words, the farther away from the target the collision happens, the greater the punishment.

\begin{equation}
rew = \left\{
\begin{array}{ll}
    rew_{dist} - 1 -\delta(2min(rew_{dist}, 0.5))^{\eta} & if\; colliding\\
    rew_{dist} -1 & not\; colliding
\end{array}
\right.
\end{equation}

where the $\delta$ and $\eta$ are hyperparameters. 

\subsection{MARA Collision Orient}
 This is the most complex environment for now. It is a combination of MARA Collision and MARA Orient, where collisions and the pose of the end-effector are taken into consideration.

\textbf{Reward system}: In the same way as in the MARA Collision environment, we punish the actions leading to collision. But in this case, we use as a core reward the one coming from the MARA Orient:

\begin{equation}
rew = \left\{
\begin{array}{ll}
    rew_{core} - \delta(2rew_{dist})^{0.03} & if\; colliding\\
    rew_{core} & not\; colliding
\end{array}
\right.
\end{equation}

\section{Experiments and results}

Since the validity of the first version of gym-gazebo was already evaluated for research purposes \cite{vilches2018robot_gym, kojcev2018hierarchical, tongloy2017asynchronous, yoonpath, zenglearning, sarkar2018sequential, zhang2018danger}, in this work we focus on the more ambitious task of achieving the first optimal policies via self learning for a robotic arm enabled by ROS 2. At Acutronic Robotics we keep pushing the state-of-the-Art in robotics and RL by offering our simulation framework as an open source project to the community, which we hope will help advancing this challenging and novel field. We also provide initial benchmarking results for different environments of the MARA robot and elaborate our results in this work. 

The experiments relying on gym-gazebo2 environments will be located in ROS2Learn \footnote{\url{http://github.com/acutronicrobotics/ros2learn}}, which contains a collection of algorithms and experiments that we are also open sourcing. You will find only a brief summary of our results below; for more information please take a look at ROS2Learn, which has been released with its own white-paper, where a more in-depth analysis of the achieved results is presented \cite{1903.06282}.
% \todo{reference to ROS2Learn paper.}

\textbf{Result summary}
Proximal Policy Optimization (PPO) \cite{schulman2017proximal} has been the first algorithm with which we have succeeded to learn an optimal policy. As aforementioned, our goal is to reach a point in the space with the end-effector of MARA. We want to reach that point with high accuracy and repeatability, which will further pave the way towards more complex tasks, such as object manipulation and grasping.

In this work, we present results trained with a MLP (Multilayer Perceptron) neural network. In the experiment the agent learns to reach a point in space by repetition. Learning to strive for high reward action-space combinations takes a long time to fully converge, as we are aiming for a tolerable error in the range of few millimeters. Using this trial and error approach the MARA training takes several hours to learn the optimal trajectory to reach the target. During our training, we have noticed that reaching the point is quite fast, but in order to have good accuracy, we need to train longer. Note that these are initial experiments of the new architecture, therefore there are many further improvements that could be made in order to achieve faster convergence.

Different problems will require different reward systems. We have developed collision and end-effector orientation aware environments, which will help us mold the policies and adapt them to our needs.

We observe a similar pattern during the training across environments. The robot reaches the target area within a range of 10-15 cm error in few hundred steps. Once the agent is at this stage it will start to reduce the distance target in a slower manner. Note that the more difficult the point is to reach (e.g. agent affected by near obstacles), the longer it will take to fully converge. Once MARA achieves consistently a few millimeters error, we consider that the policy has converged, see Figure \ref{fig:converged} for details.

\begin{center}
\label{fig:converged}
\captionsetup{type=figure}
  \includegraphics[height=70mm]{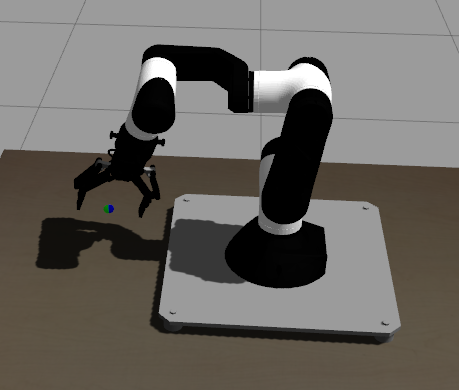}
  \caption{Solved environment, the target is reached.} 
\end{center}

\subsection{Converged Experiments}
We present converged results for each one of the environments we are publishing. We obtained the accuracy values by computing the average results (mean and standard deviation) of 10 runs from the trained neural network with PPO algorithm. Due to the stochastic nature in the run of them, we applied a stop condition in order to reliably measure the accuracy. 

The reward system is slightly tuned in some cases, as mentioned before, but our goal is to unify the reward system in a common function once we find the best hyperparameters. Again, note that there is still room for improvement for each training, so we expect to achieve faster training time and better results in the near future.

Description of all environments is available at Section \ref{environments}.

\subsubsection{MARA}
It is not expected that an environment where collisions are not considered to compute the reward learns to avoid collisions when the target is close to an object. It could also happen that in order to reach some position the agent collides with itself slightly. This environment sets the base for more complex environments, where the only parameter we need to optimize in the reward is the distance between the end-effector and the desired target.

\begin{table}[h]
\centering

\resizebox{\columnwidth}{!}{%
\begin{tabular}{|c|c|c|c|}
\hline
 \multirow{2}{*}{Accuracy} & \multicolumn{3}{|c|}{Axis} \\ \cline{2-4}
 
  & \emph{x} & \emph{y} & \emph{z} \\
  \hline
  Distance(mm) & $5.74\pm6.73$&$6.78\pm5.27$&$6.38\pm4.72$\\
  \hline
\end{tabular}
}
\caption{Mean error distribution of the training with respect to the target for $MARA$ environment}
\label{table:allalgs1}
\end{table}

\begin{figure}[h]
\centering
\captionsetup{type=figure}
  \includegraphics[width=0.48\textwidth]{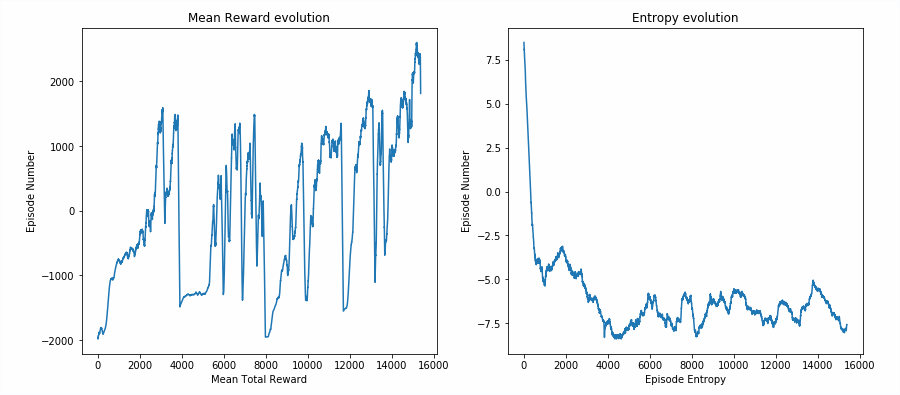}
  \caption{Mean reward evolution and entropy evolution using $MARA$ environment in the training} 
\end{figure}

\subsubsection{MARACollision}
Having the goal of transferring the policy to a real robot, we must ensure that we achieve a collision-free behaviour. By penalizing collisions we completely avoid this issue. Taking collisions into consideration during training is essential in order to represent more realistic scenarios that could occur in industrial environments.

\begin{table}[h]
\centering

\resizebox{\columnwidth}{!}{%
\begin{tabular}{|c|c|c|c|}
\hline
 \multirow{2}{*}{Accuracy} & \multicolumn{3}{|c|}{Axis} \\ \cline{2-4}
  & \emph{x} & \emph{y} & \emph{z} \\
  \hline
  Distance(mm) & $5.68\pm7.34$&$4.53\pm6.61$&$3.13\pm7.85$\\
  \hline
\end{tabular}
}
\caption{Mean error distribution of the training with respect to the target for $MARACollision$ environment}
\label{table:allalgs2}
\end{table}

\begin{figure}[h]
\centering
\captionsetup{type=figure}
  \includegraphics[width=0.48\textwidth]{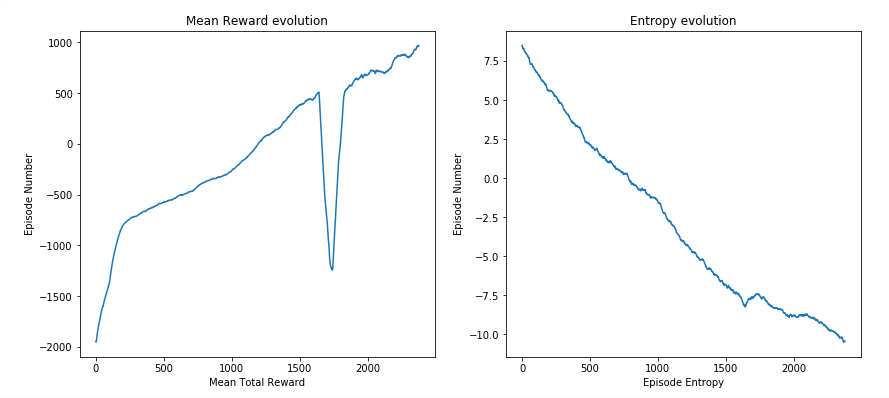}
  \caption{Mean reward evolution and entropy evolution using $MARAOrient$ environment in the training} 
\end{figure}

\subsubsection{MARAOrient}
In many real world applications it is required that the end-effector has a specific orientation, e.g. peg placement, finishing, painting. The goal is therefore to balance the trade-off between rewarding distance and orientation. The already mentioned peg placement task, for instance, would not admit any distance or pose error, which means that the reward system should be shaped to meet this requirement. For a task where we admit a wider error in the orientation, e.g pick and place, rewarding the distance to the target higher than the orientation would result in a faster learning. In the following experiment, we try to balance the desired distance and orientation, as we try to achieve good results in both aspects. We have chosen to train the end-effector to look down, which is replicating a pick and place operation. Keep in mind that, if the orientation target is much more complex, the reward system and the neural architecture might require hyperparameter optimization.

Reward system modification: $\beta$ = 1.1. As part of the search for optimal hyperparameters, this evaluation was performed with a small variation in $\beta$. This value affects the contribution of the orientation to reward function; for higher values, the system will be more tolerant to deviations in the orientation, while lower $\beta$ values will impose a higher penalty to the total reward, as Eq.\ref{eq:rew_core} shows.

\begin{table}[h]
\centering

\resizebox{\columnwidth}{!}{%
\begin{tabular}{|c|c|c|c|}
\hline
 \multirow{2}{*}{Accuracy} & \multicolumn{3}{|c|}{Axis} \\ \cline{2-4}
 
  & \emph{x} & \emph{y} & \emph{z} \\
  \hline
  Distance(mm) & $3.03\pm1.89$&$8.95\pm2.54$&$5.85\pm4.38$\\
  \hline
  Orientation (deg) & $0.71\pm0.29$&$1.61\pm1.25$&$7.52\pm 2.48$\\
  \hline
\end{tabular}
}
\caption{Mean error distribution of the training with respect to the target for $MARAOrient$ environment}
\label{table:allalgs3}
\end{table}

\begin{figure}[h]
\centering
\captionsetup{type=figure}
  \includegraphics[width=0.48\textwidth]{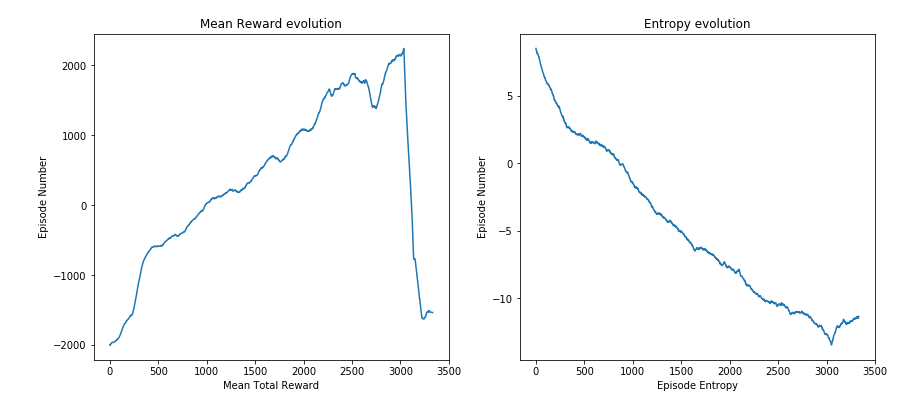}
  \caption{Mean reward evolution and entropy evolution using $MARAOrient$ environment in the training} 
\end{figure}

\subsubsection{MARACollisionOrient}
Once again, we introduce collision in the reward system in order to avoid possible unwanted states.

Reward system modification: Beta = 1.5. Once again we play with different hyperparameters trying to reach the optimal policy.

\begin{table}[h]
\centering

\resizebox{\columnwidth}{!}{%
\begin{tabular}{|c|c|c|c|}
\hline
 \multirow{2}{*}{Accuracy} & \multicolumn{3}{|c|}{Axis} \\ \cline{2-4}
 
  & \emph{x} & \emph{y} & \emph{z} \\
  \hline
  Distance (mm) & $7.43\pm3.07$&$4.69\pm2.37$&$5.18\pm3.44$\\
  \hline
  Orientation (deg) & $2.62\pm3.80$&$4.06\pm2.20$&$6.43\pm6.81$\\
  \hline
\end{tabular}
}
\caption{Mean error distribution of the training with respect to the target for $MARACollisionOrient$ environment}
\label{table:allalgs4}
\end{table}

\begin{figure}[h]
\centering
\captionsetup{type=figure}
  \includegraphics[width=0.48\textwidth]{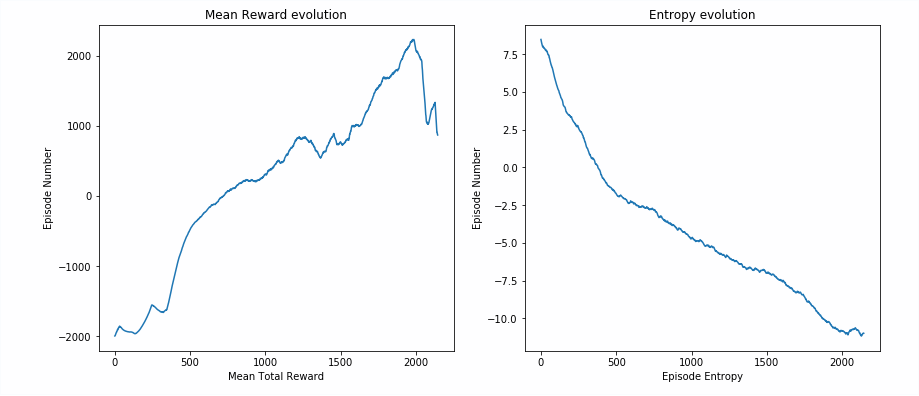}
  \caption{Mean reward evolution and entropy evolution using $MARACollisionOrient$ environment in the training} 
\end{figure}

\section*{Conclusion and Future work}
In this work, we presented an upgraded version gym-gazebo toolkit for developing and comparing RL algorithms using ROS 2 and Gazebo. We have succeded in porting our previous work \cite{zamora2016extending} to ROS 2. The evaluation results show that DRL methods have good accuracy and repeatability in the range of a few millimeters. Even tough faster convergence and algorithm stability can be reached with hyperparameter tuning, our simulation framework shows the feasibility of training modular robots leveraging the state of the art robotics tools and RL algorithms.

We plan to further extend and maintain gym-gazebo2, including different types of modular robots configurations and components. For example, we plan to add support for different type of grippers such as vacuum gripper or flexible gripper. Including different types of sensors such as force/torque sensor is also important area we would like to explore and extend their support in our environments. Another important aspect would be incorporating domain randomization features \cite{tobin2017domain, peng2018sim}, which can be used in combination with Recurrent Neural Networks (RNNs) \cite{lipton2015critical, jozefowicz2015empirical}. This will further enhance the environments to support learning and adapt their behavior with respect to environment changes. For instance, visual servoing, dynamic collision detection, force control or advanced human robot collaboration.

We expect feedback and contributions from the community and will be giving advice and guidance via GitHub issues. All in all, we highly encourage the community to contribute to this project, that will be actively maintained and developed.

\appendix
\label{appendix}

\begin{table}[h]
    \caption{Values of the used hyperparameters in the experiments} 
    \label{tab:hyperparameters}
    \begin{center}
        \begin{tabular}{|l|c|}
            \hline
                \multirow{1.5}{*}{Hyperparameter} & \multirow{1.5}{*}{Value} \\[1ex]
            \hline
                number of layers & 2 \\
            \hline
                size of hidden layers & 64 \\
            \hline
                layer normalization & False \\
            \hline
                number of steps & 2048 \\
            \hline
                number of  minibatches& 32 \\
            \hline
                lam & 0.95 \\
            \hline
                gamma & 0.99 \\
            \hline
                number of epochs & 10 \\
            \hline
                entropy coefficient & 0.0 \\
            \hline
                learning rate & $3e^{-4}*(1-\frac{current_timestep}{total_timesteps})$ \\
            \hline
                clipping range & 0.2 \\
            \hline
                value function coefficient & 0.5 \\
            \hline
                seed & 0 \\
            \hline
                value network & 'copy' \\
            \hline
                total timesteps & $1e^{8}$ \\
            \hline
        \end{tabular}
    \end{center}
\end{table}

\textit{Note: In MARACollision we used 1024 steps instead of 2048 to achieve the convergence}

\bibliographystyle{IEEEtran}
\bibliography{references}

\newpage
\onecolumn

% include the release notes page
%\input{gym_gazebo.tex}

\end{document}